\documentclass{article}

\usepackage{graphicx}

\usepackage[preprint]{neurips_2024}

\usepackage[utf8]{inputenc} 
\usepackage[T1]{fontenc}    
\usepackage{hyperref}       
\usepackage{url}            
\usepackage{booktabs}       
\usepackage{amsfonts}       
\usepackage{nicefrac}       
\usepackage{microtype}      
\usepackage{xcolor}         
\usepackage{multirow}
\usepackage{subcaption}

\usepackage{tikz}
\newcommand{\circled}[1]{%
  \begin{tikzpicture}[baseline=(char.base)]
    \node[shape=circle,draw,inner sep=1pt, line width=1pt] (char) {#1};
  \end{tikzpicture}%
}

\newcommand{\NAME}{FedRouter}

\title{Task-Centric Personalized Federated Fine-Tuning of Language Models}

\author{
 \textbf{Gabriel U. Talasso}\textsuperscript{1}\thanks{Correspondence: g235078@dac.unicamp.br} \quad
  \textbf{Meghdad Kurmanji} \textsuperscript{2} \quad
  \textbf{Allan {M. de Souza}}\textsuperscript{1} \quad  \\
  \textbf{Nicholas D. Lane} \textsuperscript{2,3} \quad
  \textbf{Leandro A. Villas} \textsuperscript{1} \\
  \textsuperscript{1}\, Universidade Estadual de Campinas \\
\textsuperscript{2}\, University of Cambridge \\
\textsuperscript{3}\, Flower Labs
}

\begin{document}
\maketitle

\begin{abstract}
Federated Learning (FL) has emerged as a promising technique for training language models on distributed and private datasets of diverse tasks. However, aggregating models trained on heterogeneous tasks often degrades the overall performance of individual clients. To address this issue, Personalized FL (pFL) aims to create models tailored for each client’s data distribution. Although these approaches improve local performance, they usually lack robustness in two aspects: (i) generalization: when clients must make predictions on unseen tasks, or face changes in their data distributions, and (ii) intra-client tasks interference: when a single client's data contains multiple distributions that may interfere with each other during local training. To tackle these two challenges, we propose \NAME, a clustering-based pFL that builds specialized models for each task rather than for each client. {\NAME} uses adapters to personalize models by employing two clustering mechanisms to associate adapters with specific tasks. A local clustering that associate adapters with task data samples and a global one that associates similar adapters from different clients to construct task-centric personalized models. Additionally, we propose an evaluation router mechanism that routes test samples to the best adapter based on the created clusters. Experiments comparing our method with existing approaches across a multitask dataset, {\NAME} demonstrate strong resilience in these challenging scenarios performing up to \textbf{$\sim$6.1\%} relatively better under tasks interference and up to \textbf{$\sim$136\%} relative improvement under generalization evaluation.
\end{abstract}

\section{Introduction}
\label{sec:intro}


Foundation models have gained significant attention in recent years, particularly due to their remarkable ability to be applied across diverse domains. In particular, Large Language Models (LLMs) have been successfully employed in a wide range of applications such as mobile devices, healthcare, and law \cite{FedFM}. To achieve strong performance in these domains, LLMs are fine-tuned, in a post-training process, for specific tasks to enhance their specialization and effectiveness \cite{lora,hydralora}. However, this adaptation process requires access to large amounts of high-quality, domain-specific data to properly align the model’s behavior with the target task \cite{openfedllm}. In this context, Federated Learning (FL) emerges as a promising paradigm for adapting foundation models, as it enables access to distributed, high-quality datasets located across multiple clients while preserving data privacy by sharing only model parameters rather than raw data \cite{FedFM,openfedllm,worldwide,sani2025photon}.


Since fine-tuning large-scale models is expensive, Parameter-Efficient Fine-Tuning (PEFT) \cite{peft} methods, such as the Low-Rank Adapters (LoRA) \cite{lora}, have been explored to significantly reduce resource demands and have also been adapted to FL scenarios~\cite{openfedllm,worldwide,fedit,fedscope}. A second major challenge arises from the data heterogeneity across clients in FL, where datasets may differ in domains, distributions, or even underlying tasks. This variability often leads to degraded performance when aggregating locally trained models into a unique global one. To mitigate this issue, state-of-the-art approaches focus on personalized FL (pFL), aiming to produce tailored models for each client \cite{ffa-lora,FedSA,feddpa}. These methods seek to \emph{balance collaborative knowledge sharing with client-specific adaptation}, thereby improving overall performance in heterogeneous scenarios~\cite{smith2017federated}.


\paragraph{Personalized FL Challenges.} Client-specific models introduce several challenges, including generalization and intra-client tasks interference, that we aim to address in this paper. First is \textbf{generalization}. On the one hand, pFL methods tailor models for each client; changes in local data distributions, such as the rise of new tasks at test time, can lead to significant performance degradation, as the models were not optimized for these scenarios \cite{feddpa}. On the other hand, \textbf{tasks interference} arises from divergent tasks within the same client’s dataset. In such cases, training a single adapter per client is suboptimal, as it must fit to potentially conflicting objectives, similar to multi-task learning techniques \cite{crawshaw2020multi,hydralora}. This negative interference~\cite{crawshaw2020multi} often degrades adaptation, consequently reducing overall performance across heterogeneous client datasets.

\begin{figure*}[t]
    \centering
    \includegraphics[width=1\textwidth]{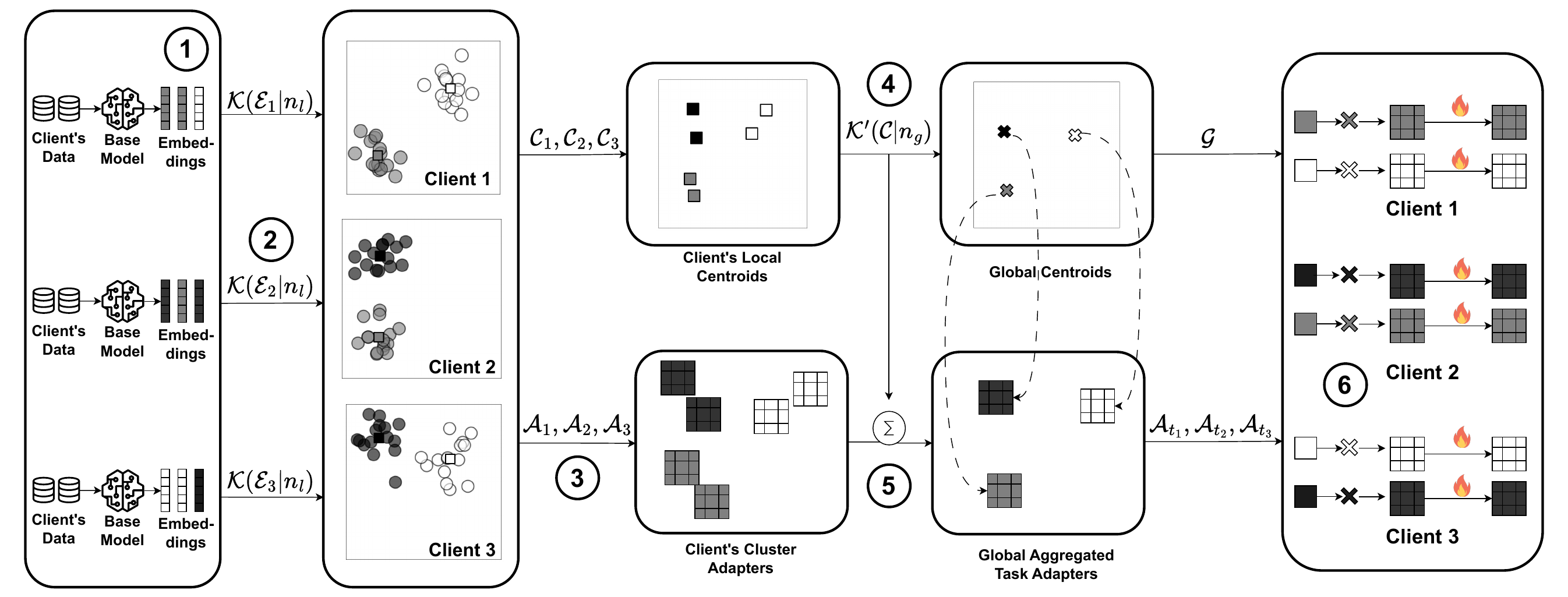}
    \caption{\textbf{\NAME~Workflow Overview.} Each client first computes embeddings from its local data and applies clustering to partition the dataset into task-specific subsets. The client then sends the resulting centroids and adapters to the server, which performs global clustering to associate similar tasks across clients and aggregate their corresponding adapters collaboratively. Finally, the server sends the updated adapters back to the clients, which then associate each received model with the appropriate local task-specific dataset for next round of training.}
    \label{fig:overview}
\end{figure*}


\paragraph{Contributions.} We propose \textbf{\NAME}, a clustering-based pFL method for federated adaptation of LLMs across multiple tasks. Our approach shifts the personalization perspective from a \emph{client-centric} paradigm, where individual models are tailored for each client, to a \emph{task-centric} one, where specialized models are created for each task. This design mitigates issues caused by both generalization failures and tasks interference. \textbf{During training}, {\NAME} leverages two complementary clustering methods: \textit{(i)} a local clustering process that partitions each client’s data into distinct tasks and trains a specialized adapter for each of them to avoid divergences issues; and \textit{(ii)} a global clustering process that groups similar tasks across clients. \textbf{During inference}, we introduce an adaptive router evaluation mechanism with local and global modes. In the local mode, new data samples are routed to the most relevant local task clusters, ensuring personalized inference. In the global mode, samples can be matched to any task cluster present across the federation, enabling generalized inference even under test-time distribution shifts or the presence of previously unseen tasks. Our main contributions are summarized as follows:


\begin{itemize}
\item We first identify and analyze two major challenges within pFL settings: $(i)$ the generalization problem under test-time distribution shifts such as the introduction of new tasks in client test datasets, and $(ii)$ the negative impact of intra-client tasks interference, when clients possess multiple, heterogeneous datasets leading to conflicting optimization objectives.
\item We propose \NAME, a task-centric pFL method that leverages both local and global clustering mechanisms to collaboratively train specialized models for each task, effectively mitigating generalization issues and intra-client tasks interference. And made the implementation publicly available\footnote{https://github.com/GabrielTalasso/FedRouter}.
\item We propose a two-mode adaptive inference pipeline, where new samples are dynamically routed either to local adapters for personalization or to the full adapter pool for global generalization, enabling unified personalized and generalized evaluation at test time.
\item We conduct an extensive empirical evaluation of {\NAME} under the identified challenging scenarios, demonstrating its superior performance compared to traditional client-centric personalization approaches, performing until 3.5\% ($\sim$6.1\% relative) better under tasks interference and until 33.6\% ($\sim$136\% relative) improvement under generalization evaluation.
\end{itemize}

\section{Related Works}

\textbf{Federated Fine-Tuning of Language Models.} With the evolution of current language models, there is an increasing need for large amounts of high-quality data for training. In this context, several studies use FL as an approach to access more training data while preserving user privacy \cite{pmlr-v54-mcmahan17a,worldwide}. In particular, works such as FedIT \cite{fedit}, OpenFedLLM \cite{openfedllm} and FederatedScope-LLM \cite{fedscope} apply FL for post-training of these models, especially by leveraging Parameter-Efficient Fine-Tuning (PEFT) \cite{peft,lora} techniques to improve training efficiency and reduce communication costs. These studies highlight the limitations related to fine-tuning models across multiple domains and tasks, claiming the need for new personalization approaches tailored for these scenarios \cite{survey_pfl}.

\textbf{Personalized Federated Language Models.} Based on the limitations of fine-tuning language models using FL in heterogeneous tasks and domains, some solutions address this problem by creating personalized models for each client or group of clients. The first class of works focuses on creating models for clusters of similar clients by measuring their similarity and aggregating models separately, thus avoiding the negative interference caused by different data distributions \cite{cfl,ifca,fedsccs,fedmllm}. Other approaches personalize the model by dividing the total training parameters into two groups: one to learn the general knowledge shared across the federation and another to learn the specific knowledge of each client. For example, FedDPA \cite{feddpa} proposes a globally shared adapter trained individually on each client, along with a local adapter trained with consideration of the global one to adapt the model to local task distributions. On the other hand, FFA-LoRA \cite{ffa-lora} and FedSA \cite{FedSA} train part of an adapter locally (the B matrix of LoRA in both cases) while either freezing (in FFA-LoRA) or sharing (in FedSA) the A matrix, which represents the general knowledge.

Although these solutions represent advances in training for multiple tasks in a federated manner, several challenges remain for such approaches. The first concerns generalization: although in some cases clients share a common component (as in FedDPA and FedSA), changes in data distributions and the addition of new tasks lead to a significant drop in the performance of models that are specific to each client. Another challenge involves the existence of multiple domains or tasks locally, which hinders model optimization even before sharing and aggregation step. These open challenges are the focus of our proposal in this work.

\section{\NAME}

We propose {\NAME}, a task-centric personalization approach that leverages clustering to create specialized models in FL. {\NAME} improves both \emph{generalization} and \emph{tasks interference} challenges in the presence of heterogeneous data and unseen tasks and domains. Our proposed approach is composed of three main components (i) Local clustering of raw data (ii) Global clustering of task centroids and (iii) Evaluation Router Mechanism. Additionally, each of these components is composed of steps that are described in Figure~\ref{fig:overview} for clustering mechanisms and Figure~\ref{fig:eval} for routing details.

\subsection{Local Clustering Mechanism}

Data can be produced by various applications and sources on the client's side. Consequently, it is not reasonable to assume that all clients have their data separated by task in the same way. Furthermore, a single task can be subdivided into several others or grouped with similar tasks to improve model performance. As a result, clients’ data may be mixed, and the definition of tasks and subtasks can become ambiguous. To address this problem, \NAME~firstly aims to separate different tasks within each client’s local dataset in order to train specialized adapters for each of them.

To this end, the first step \circled{1} involves computing the embeddings of the training data using the pre-trained base model. This step is performed only once as a prerequisite to starting the federation and results in a set of embeddings ($\mathcal{E}_i$) for client $i$, with dimensions $D$ (dataset length) by $E$ (model-dependent embedding size). Next, in the second step \circled{2}, clients perform local clustering on the set $\mathcal{E}_i$ using the clustering algorithm $\mathcal{K}(\mathcal{E}_i|n_l)$, where $n_l$ denotes the number of local clusters to be created. In our case, we set $\mathcal{K}$ to K-Means, without loss of generality, since other methods capable of generating a centroid representing each cluster could also be used. Additionally, the number of clusters may vary across clients depending on the degree of local task heterogeneity. More details about these hyperparameters are provided in the results section.

Finally, once the embeddings have been computed and local clusters created, the third step \circled{3} involves training a specialized adapter for each cluster and sharing the corresponding parameters $\mathcal{A}_i$ and the set of centroids $\mathcal{C}_i$ with the server. To avoid excessive communication and computation costs associated with training and transmitting multiple models, we propose a round-robin-based approach, in which only one cluster is trained per round and its adapter is shared. The coordination of which adapter will be trained is made by the server. Thus, in later rounds, each client shares only one centroid and one adapter.

\subsection{Global Clustering Mechanism}
\label{subsec-globalcluster}

As the clients may have similar tasks of other clients, we need a way to associate the same cluster of tasks with different clients. For this, the second main component of {\NAME} is the server-side clustering mechanism, which aims to associate similar tasks across different clients while aggregating the corresponding models for each task. 

In the fourth step \circled{4}, during the first communication round, all clients share their local centroids, which serve as proxies for each client’s task data. The server then performs another clustering over all shared centroids, denoted as $\mathcal{K}(\mathcal{C}|n_g)$, where $\mathcal{C} = \mathcal{C}_1 \cup \mathcal{C}_2 \cup \dots \cup \mathcal{C}_N$ and $n_g$ represents the number of global clusters, corresponding to the total number of tasks in the federation. Using the resulting global centroids $\mathcal{G}$ for each task, the server aggregates the associated adapters through averaging in step \circled{5}. As in the local phase, we employ K-Means as the clustering algorithm, but other clustering methods could be used without loss of generality. Likewise, while we use averaging for aggregation, other aggregation strategies could also be applied.

To avoid broadcasting all adapters to clients in every round, the server selects only the next centroid and its corresponding adapter to be sent in a round-robin-based coordination, i.e, each client will train only one adapter of one task per round, avoiding increasing computer and communication requirements. In a later round, the client may receive other adapters and tasks to be trained. This process is managed by the server, which sends the correct centroids to each client train. Finally, in step \circled{6}, each client associates the received global centroid with its corresponding local centroid by computing the Euclidean distance between them, and then retrains the associated adapter locally using the data from that local cluster. This cycle is repeated for $T$ rounds or until a defined stopping criterion.

\subsection{Evaluation Router Mechanism}
\label{sec:eval_router}

\begin{figure*}[hbt]
    \centering
    \includegraphics[width=0.9\textwidth]{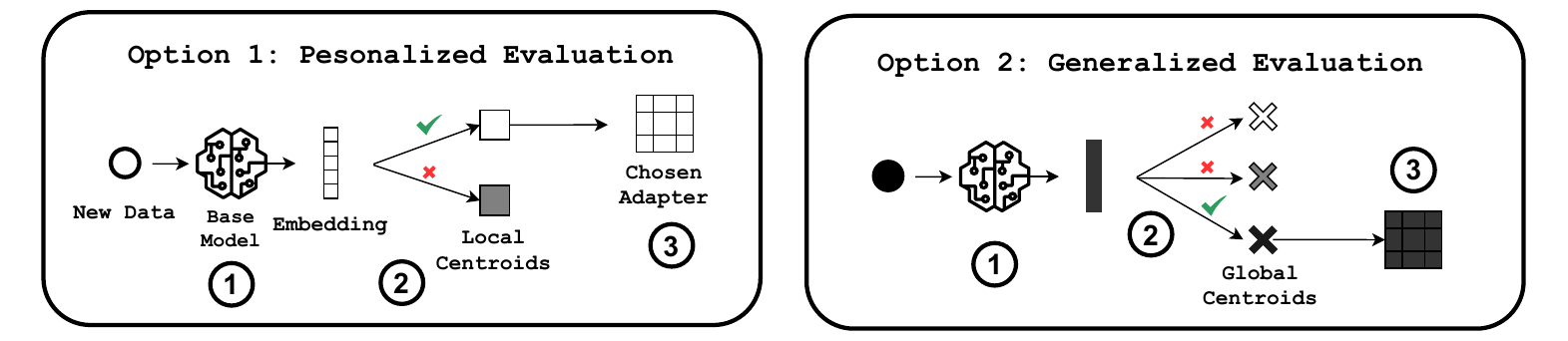}
    \caption{\textbf{\NAME~Evaluation Modes.} During inference, each client computes the embedding of a new data sample and associates it with the nearest centroid based on the minimum Euclidean distance. The association can be performed using either the \textbf{local} centroids, to obtain a personalized evaluation, or the \textbf{global} centroids, to enable a generalized evaluation across the federation.}
    \label{fig:eval}
\end{figure*}

As illustrated in Figure~\ref{fig:eval}, we introduce a novel inference mechanism, which aims to promote generalization in scenarios with test-time distribution shifts \cite{feddpa}, where clients must evaluate on data from distributions unseen during local training. Such situations occur when new tasks are introduced in the test datasets or when client data distributions change over time. To address this, \NAME~supports two possible evaluation modes, illustrated in Figure~\ref{fig:eval}, that can be chosen depending on the existence of new tasks on test-time dataset, allowing the generalization.

In the first step \circled{1} of both modes, the embeddings of the new test samples are computed only using the pre-trained base model. In the second step \circled{2}, each embedding is associated with its nearest adapter by finding the minimum Euclidean distance to the centroids, ensuring that each sample is evaluated using the most appropriate task-specific adapter. To improve efficiency when testing multiple data points, all samples are first assigned to their corresponding nearest adapter. Then, each adapter is set once, and the corresponding batches are evaluated together, thereby avoiding unnecessary adapter switching. Finally, in the third step \circled{3} the nearest chosen adapter is set to model inference process.

The difference between the two evaluation modes (local and global) lies only in the centroids used to route the new data. In the local mode, only the client’s locally computed centroids are available for association, whereas in the global mode, all global centroids are accessible. The local mode enhances routing accuracy in scenarios without test-time distribution shifts, as fewer options lead to more precise associations. Conversely, the global mode improves performance when distribution shifts occur, since it allows the use of adapters associated with previously unseen tasks.

\section{Results}

This section presents an extensive evaluation of our proposed method, \NAME, under scenarios where most existing pFL approaches struggle to perform effectively. We begin by describing the experimental setup, including the models, datasets, scenarios, and baseline methods for comparison (Section~\ref{sec:eval_setup}). Next, we report and discuss the main results, first to answer if {\NAME} improves task interference issues (Section~\ref{sec:results_interference}) and then if it improve generalization issues (Section~\ref{sec:results_interference}). Finally, we conduct ablation studies to analyze the scalability and impact of each component and hyperparameter in our framework (Section~\ref{sec:ablation}).

\subsection{Evaluation Setup}
\label{sec:eval_setup}

\textbf{Datasets.} To evaluate data heterogeneity in multitask language model fine-tuning, we select a subset of four tasks from FLAN \cite{flan}: QQP for paraphrase detection, WebNLG for structure-to-text generation, Samsum for dialogue summarization, and GigaWord for general text summarization. Following \cite{feddpa}, we use ROUGE-1 as the primary evaluation metric.

\textbf{Baselines.} To assess the effectiveness of \NAME, we compare it against several representative baselines: \textbf{FedIT} \cite{fedit} (the instruction-tuning variant of FedAvg), \textbf{Local} (a non-federated, independent fine-tuning baseline), \textbf{FedCluster} (a clustering-based method inspired by \cite{cfl,ifca} and adapted for LLM fine-tuning), \textbf{FedDPA} \cite{feddpa}, and \textbf{FedSA} \cite{FedSA}, which represent state-of-the-art personalized FL approaches. We also compare with a variant, called {\NAME*}, to computer budge flexibility to the training.

\textbf{Evaluation Scenarios.} We design three evaluation settings to capture different levels of intra-client tasks interference. In the \textbf{Single} scenario, each client has only one task, representing the standard assumption in most prior works, without significant interference. In the \textbf{Dual} scenario, each client has two distinct tasks simultaneously. Finally, in the \textbf{All} scenario, each client has all tasks, resulting in the highest degree of tasks interference.

\textbf{Model Setup.} We used Llama 3.2 with 1B parameters in the experiments. All models are instruction-tuned with a maximum sequence length of 1024 tokens. Experiments are performed on an NVIDIA A100 80GB GPU, using a batch size of 16 and a learning rate of $5\times10^{-4}$ for 10 steps each round. All LoRA adapters are configured with a rank of 8 and an $\alpha$ value of 16.

\textbf{Federated Setup.} We implement {\NAME} using the Flower framework \cite{flower} and the OpenFedLLM base code \cite{openfedllm} and made the implementation public available \footnote{https://github.com/GabrielTalasso/FedRouter}. All experiments were conducted with 8 clients participating in the federation for 25 rounds, where each pair of clients receives a similar dataset under each evaluation scenario. To emulate a realistic data-scarcity condition \cite{feddpa,pmlr-v54-mcmahan17a}, the training data for each client is limited to 600 samples, while the test set contains 300 samples per client.
\subsection{Task Interference Results}
\label{sec:results_interference}

The results, summarized by the average of 5 runs in Table~\ref{tab:fl_results}, show that {\NAME} achieves performance comparable to other methods in the Single scenario, where no tasks interference is present, demonstrating its ability to provide effective personalization similar to state-of-the-art approaches, especially when compared with no-personalization as FedIT. However, as client divergence increases in the Dual and All scenarios, where clients hold data from multiple tasks, the performance of other methods degrades significantly faster. In contrast, {\NAME} consistently outperforms competing methods, achieving the highest scores in most cases and on average across all scenarios. These results highlight the advantages of shifting from a client-centric to a task-centric personalization paradigm, where training specialized models per task, rather than per client, effectively mitigates tasks interference and enhances robustness under heterogeneous data conditions.

We also report in Table~\ref{tab:fl_results} the \textit{\NAME*}, a variant that makes the computation and communication budget of our approach more flexible. In contrast to standard FedRouter, where each client trains and communicates the same number of adapters per round, thereby normalizing the training effort at the client level as described in Section~\ref{subsec-globalcluster}, \textit{\NAME*} updates all adapters available at each client in every round. This normalization criterion moves from the client level to the adapter level budget, ensuring that each adapter receives (in the All Scenario, and proportional for the others) an equivalent amount of training across rounds. The results show that in this scenario, where there is no limitation on standardizing client resources, \NAME~achieves even greater performance, standing out more than other solutions.

\begin{table}[t]
\centering
\caption{Performance comparison (mean $\pm$ std) across different data scenarios. The last column reports the average performance across scenarios. Best results per column are highlighted in bold.}
\label{tab:results}
\begin{tabular}{lcccc}
\toprule
\textbf{Method} & \textbf{Single} & \textbf{Dual} & \textbf{All} & \textbf{Average} \\
\midrule
FedIT     & 0.546 $\pm$ 0.012 & 0.550 $\pm$ 0.009 & 0.560 $\pm$ 0.010 & 0.552 \\
Local     & 0.553 $\pm$ 0.005 & 0.525 $\pm$ 0.014 & 0.534 $\pm$ 0.005 & 0.537 \\
FedCluster   & \textbf{0.561} $\pm$ 0.008 & 0.551 $\pm$ 0.012 & 0.553 $\pm$ 0.012 & 0.555 \\
FedSA     & 0.554 $\pm$ 0.008 & 0.530 $\pm$ 0.010 & 0.531 $\pm$ 0.010 & 0.538 \\
FedDPA    & 0.556 $\pm$ 0.009 & 0.551 $\pm$ 0.013 & 0.549 $\pm$ 0.010 & 0.552 \\
\NAME & \textbf{0.561} $\pm$ 0.004 & \textbf{0.558} $\pm$ 0.019 & \textbf{0.566} $\pm$ 0.034 & \textbf{0.562} \\
\bottomrule
\textit{\NAME*} & \textit{\textbf{0.562}} $\pm$ \textit{0.013} & \textit{\textbf{0.563}} $\pm$ \textit{0.012} & \textit{\textbf{0.575}} $\pm$ \textit{0.012} & \textit{\textbf{0.567}} \\
\bottomrule
\label{tab:fl_results}
\end{tabular}
\end{table}

\subsection{Generalization Results}
\label{sec:results_generalization}

\begin{figure}[ht]
\begin{minipage}{0.40\linewidth}
        \centering
        \caption{Performance comparison (mean $\pm$ std) in the single-task training scenario, evaluated on all tasks at test time to assess generalization capability and robustness under test-time distribution shift.}
        \label{tab:results}
        \begin{tabular}{lcccc}
        \toprule
        \textbf{Method} & \textbf{Test-Time Gen.} \\
        \midrule
        FedIT     & 0.570 $\pm$ 0.013  \\
        Local     & 0.255 $\pm$ 0.006 \\
        FedCluster   & 0.252 $\pm$ 0.008 \\
        FedSA     & 0.247 $\pm$ 0.008\\
        FedDPA    & 0.461 $\pm$ 0.009  \\
        \NAME & \textbf{0.583} $\pm$ 0.005 \\
        \bottomrule
        \label{tab:fl_results_general}
        \end{tabular}
    \end{minipage}
    \hfill
    \centering
    \begin{minipage}{0.5\linewidth}
        \centering
        \includegraphics[width=\linewidth]{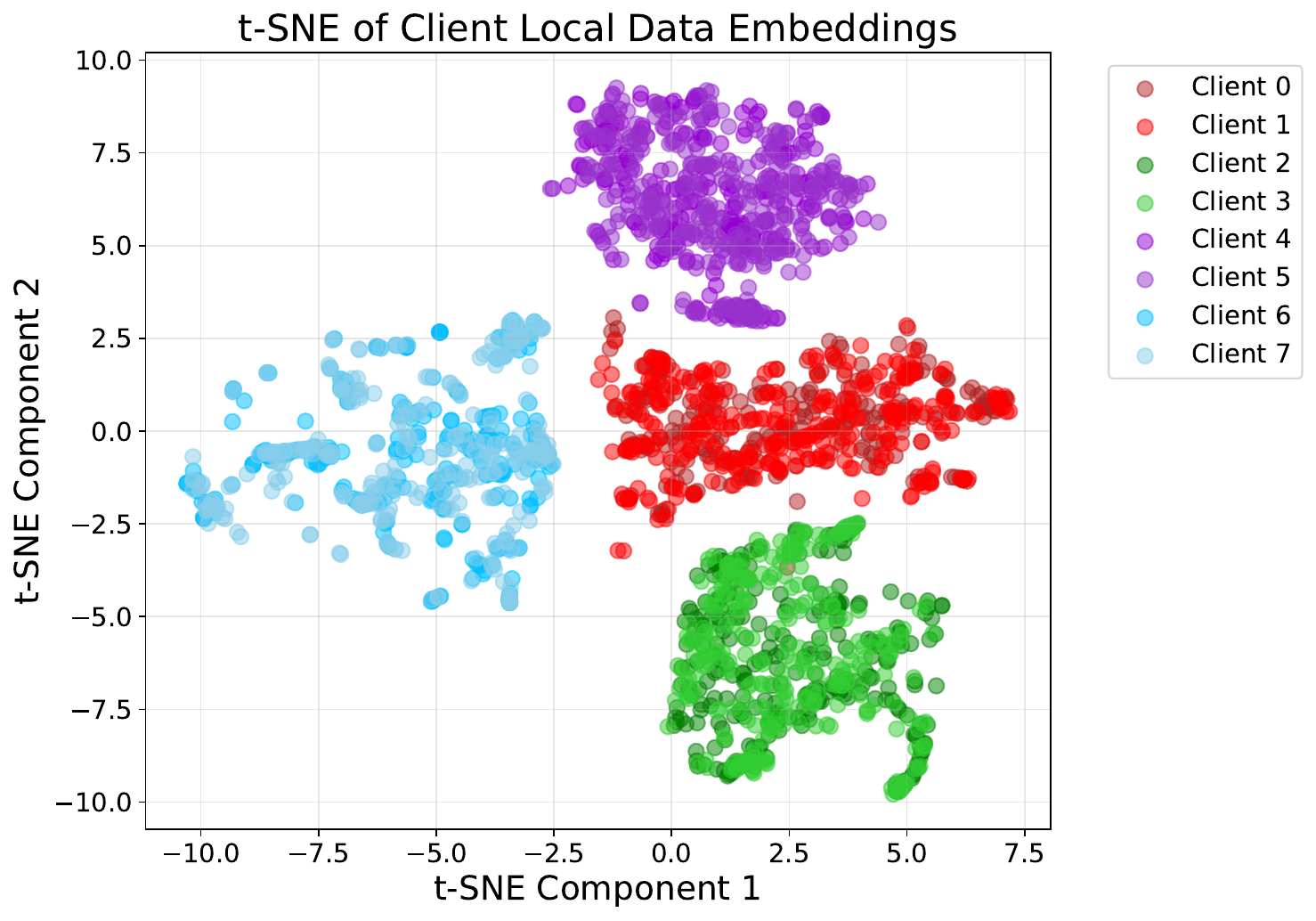}
        \caption{t-SNE visualization of client test data embeddings in single scenario.}
        \label{fig:client_emb}
    \end{minipage}
\end{figure}

To evaluate test-time distribution shift scenarios, where clients are required to perform inference on unseen tasks, we present in Table~\ref{tab:fl_results_general} the final performance of all compared methods trained on the Single scenario. 

The results reveal a substantial performance drop in most personalization-based methods, as these approaches train client-specific models that fail to generalize to unseen tasks. Two exceptions are observed: FedIT \cite{fedit}, which shows limited degradation due to aggregating updates from clients across all tasks without explicit personalization, and FedDPA \cite{feddpa}, which maintains a stable performance through its global adapter and inference mechanism designed to mitigate test-time distribution shifts. Finally, {\NAME} demonstrates the most robust generalization among all evaluated methods. By training task-specific adapters and leveraging global evaluation to reuse adapters from unseen tasks, {\NAME} achieves superior performance under distribution shifts, reinforcing the advantages of its task-centric personalization strategy in FL.

Additionally, we evaluated the quality of the local clustering method to better understand the remaining sources of error that may explain the small performance decay observed in some cases. Figure~\ref{fig:client_emb} illustrates the t-SNE of clients' local test data embeddings, in the single scenario, showing clearly separated tasks across different clients while showing that similar clients, with the same tasks, remain close together. This supports the fact that our method trains a specific adapter for each task by clustering the local data. Furthermore, the clustering accuracy evaluated on each client’s test data reached 100\%, 100\%, and 95.4\% for the Single, Dual, and All domain scenarios, respectively. These results indicate that scenarios involving a larger number of tasks are more challenging, often leading to reduced clustering accuracy due to overlapping tasks across clients.



\subsection{Ablation Studies}
\label{sec:ablation}

To fully explore {\NAME}'s behavior under considering different conditions, we performed ablation experiments varying the method and scenario to ensure its effectiveness and stability. Below, we first present experiments related to the scalability of federated scenarios, followed by experiments regarding the correct selection of the number of clusters.

\subsection{Scaling}

\begin{figure}[ht]
    \centering
    \begin{minipage}{0.46\linewidth}
        \centering
        \includegraphics[width=\linewidth]{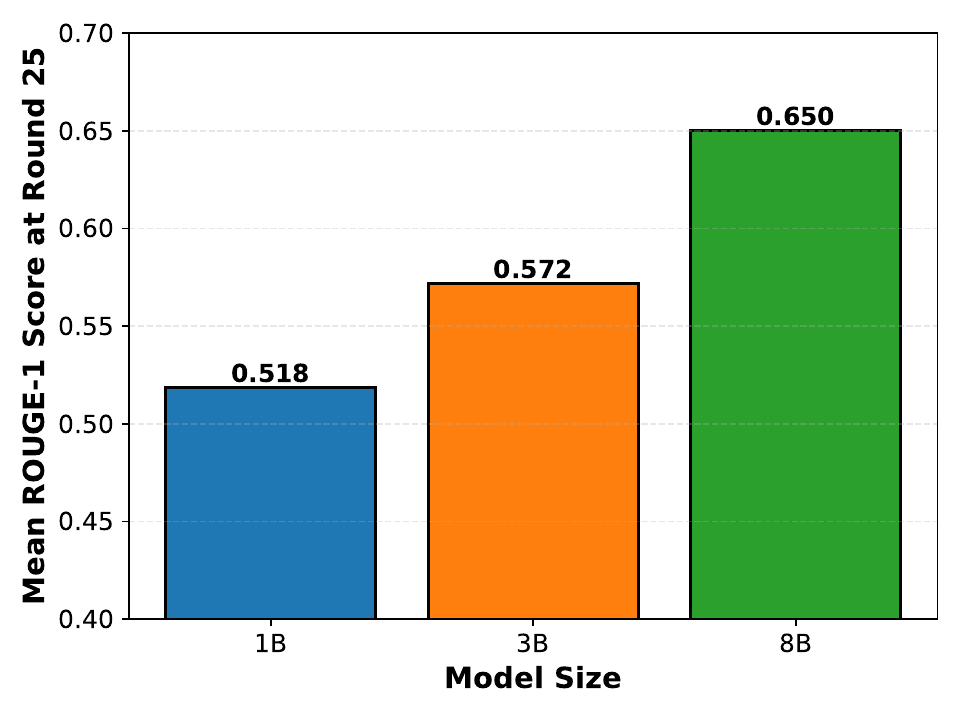}
        \caption{Scaling model size of Llama models using {\NAME} in single scenario.}
        \label{fig:model_size}
    \end{minipage}
    \hfill
    \begin{minipage}{0.46\linewidth}
        \centering
        \includegraphics[width=\linewidth]{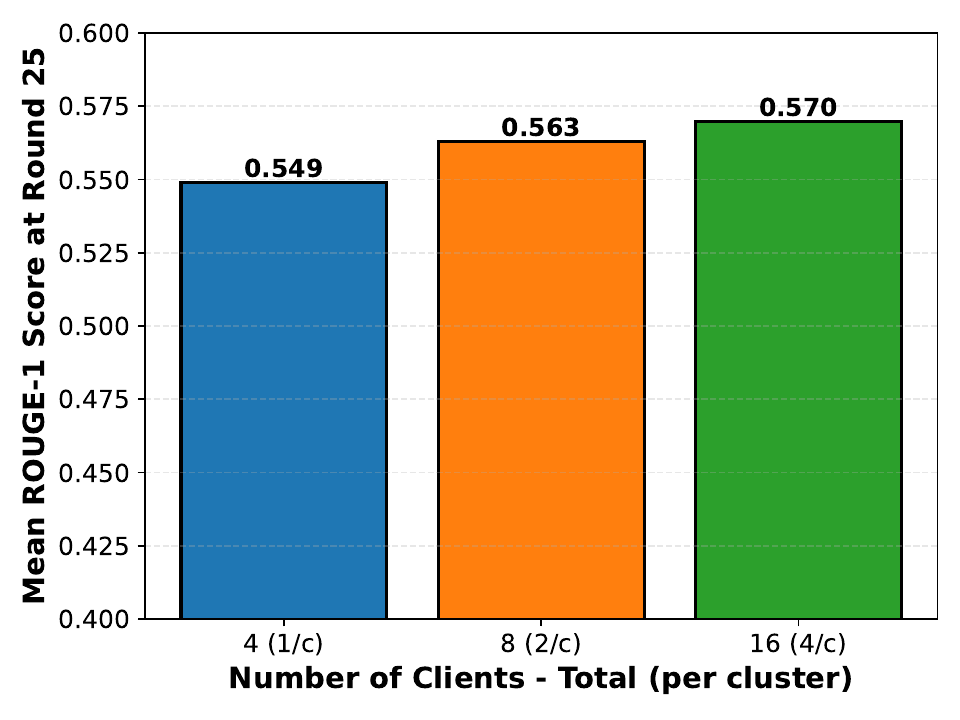}
        \caption{Scaling the number of clients, total and per cluster of \NAME~in single scenario.}
        \label{fig:number_of_clients}
    \end{minipage}
\end{figure}

Figure~\ref{fig:model_size} shows the performance of {\NAME} when scaling the model size. In this experiment, three model sizes of LLama's 3 model family (\cite{llama}), 1 billion, 3 billion, and 8 billion parameters models in the  ``single'' evaluation scenario were used, using the presented method, and limiting the batch size to 8. The results showed that {\NAME} scales the performance without a bottleneck when scaling the number of parameters, showing that our proposed method can also be used in both small and larger model scenarios and applications.

Additionally, in Figure~\ref{fig:number_of_clients} we show the results of scaling the number of clients on the federation and consequently on each cluster, also in the single scenario. The results showed that, due to the availability of more data to train the specialized adapters per task, the performance improves with more clients, which is beneficial when scaling {\NAME} to be used in scenarios with more users.

\subsection{Number of Clusters}

\begin{figure*}[hbt]
    \centering
    \includegraphics[width=0.95\textwidth]{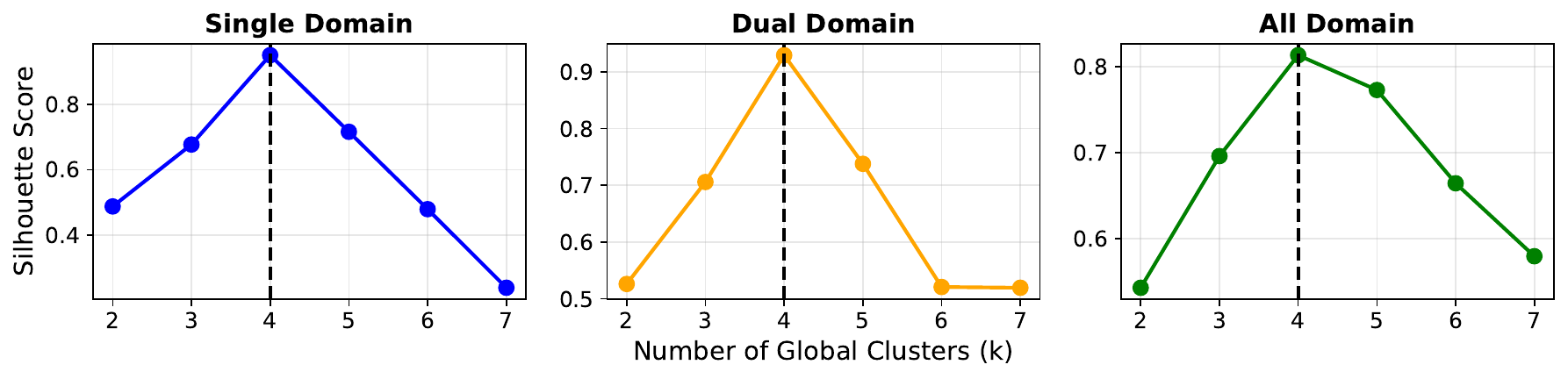}
    \caption{Silhouette Score method to choose the number of global clusters in different scenarios for \NAME.}
    \label{fig:eval_n_clusters_global}
\end{figure*}

As the number of clusters is a hyperparameter of {\NAME} in both the local clustering process and the global clustering process, we also analyze methods to choose the correct number of clusters. Figure~\ref{fig:eval_n_clusters_global} shows the results of the Silhouette Score method to find the correct number of global clusters based on the centroids reported by the clients on each of the three proposed scenarios. The results show that in all cases it is possible to clearly find the correct number of clusters, in this case 4, due the good separability of the embeddings tasks as shown previously. It results in improved performance of \NAME, as it can effectively cluster similar tasks from different clients while avoiding the aggregation of different tasks into the same adapter.

\begin{figure*}[hbt]
    \centering
    \includegraphics[width=0.65\textwidth]{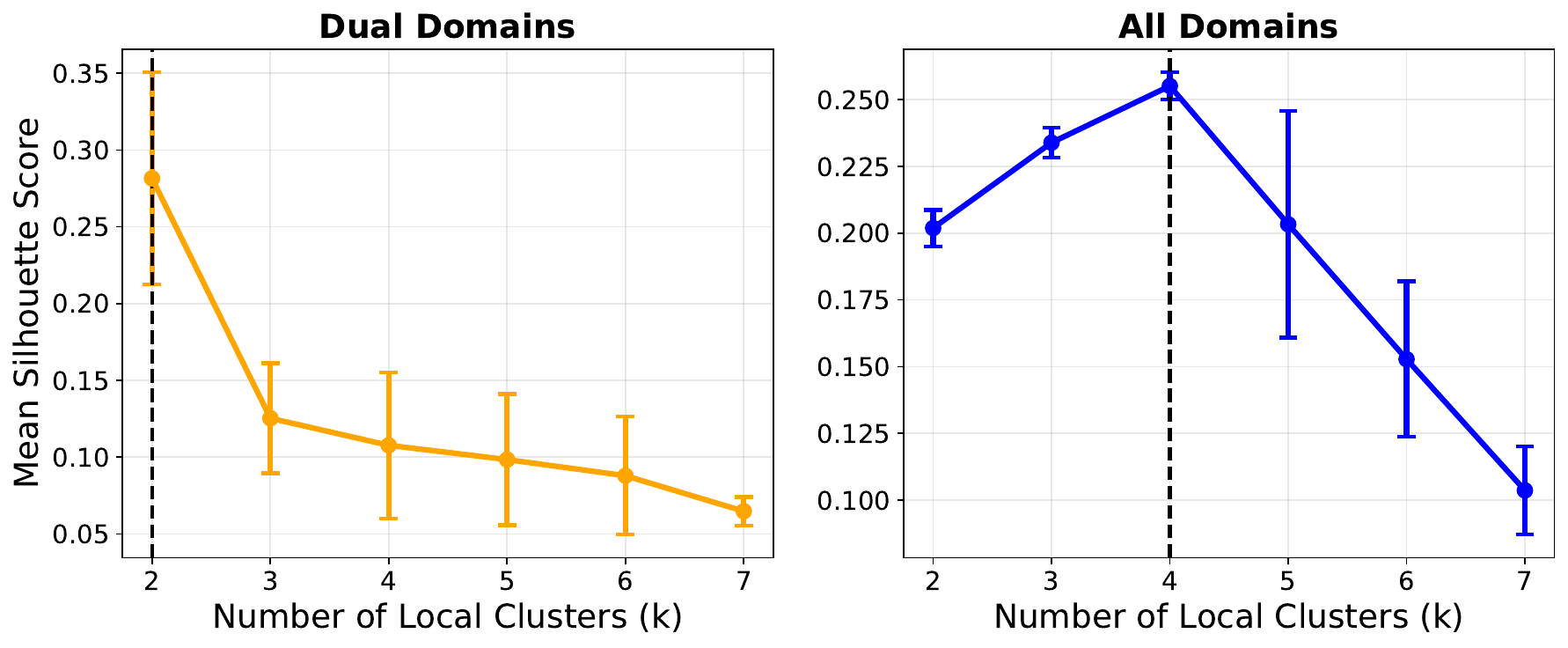}
    \caption{Silhouette Score method to choose the number of local clusters in different scenarios for \NAME.}
    \label{fig:eval_n_clusters_local}
\end{figure*}

Additionally, Figure~\ref{fig:eval_n_clusters_local} presents the results of applying the Silhouette Score method to find the number of local clusters in the Dual scenario, where the correct number is 2 tasks per client and the all scenario where the correct number is 4, in both cases the method was capable of clearly finding the best number of clusters. We do not consider the single scenario because each client has only one cluster locally, and Silhouette Score is not applied. Confidence bars represent the standard deviation due the measure on different clients' local datasets.

\section{Conclusion}

Personalization in FL fine-tuning is promising in heterogeneous and multi-task scenarios, but existing methods often ignore negative interference among tasks within clients and performance degradation under test-time distribution shifts. In this work, we proposed \NAME, a task-centric personalization framework that combines both local and global clustering mechanisms to train specialized adapters. By structuring adapters around task clusters and enabling routing at inference time, {\NAME} mitigates negative transfer and improves test-time generalization. Future work involves exploring scenarios with even more tasks on the federation and the possible cross-task collaboration methods.

\bibliographystyle{unsrtnat}
\bibliography{references}

@article{worldwide,
  title={Worldwide federated training of language models},
  author={Iacob, Alex and Sani, Lorenzo and Marino, Bill and Aleksandrov, Preslav and Shen, William F and Lane, Nicholas Donald},
  journal={arXiv preprint arXiv:2405.14446},
  year={2024}
}

@inproceedings{fedit,
  title={Towards building the federatedgpt: Federated instruction tuning},
  author={Zhang, Jianyi and Vahidian, Saeed and Kuo, Martin and Li, Chunyuan and Zhang, Ruiyi and Yu, Tong and Wang, Guoyin and Chen, Yiran},
  booktitle={ICASSP 2024-2024 IEEE International Conference on Acoustics, Speech and Signal Processing (ICASSP)},
  pages={6915--6919},
  year={2024},
  organization={IEEE}
}

@article{lora,
  title={Lora: Low-rank adaptation of large language models.},
  author={Hu, Edward J and Shen, Yelong and Wallis, Phillip and Allen-Zhu, Zeyuan and Li, Yuanzhi and Wang, Shean and Wang, Lu and Chen, Weizhu and others},
  journal={ICLR},
  volume={1},
  number={2},
  pages={3},
  year={2022}
}

@inproceedings{openfedllm,
  title={Openfedllm: Training large language models on decentralized private data via federated learning},
  author={Ye, Rui and Wang, Wenhao and Chai, Jingyi and Li, Dihan and Li, Zexi and Xu, Yinda and Du, Yaxin and Wang, Yanfeng and Chen, Siheng},
  booktitle={Proceedings of the 30th ACM SIGKDD conference on knowledge discovery and data mining},
  pages={6137--6147},
  year={2024}
}

@article{cfl,
  title={Clustered federated learning: Model-agnostic distributed multitask optimization under privacy constraints},
  author={Sattler, Felix and M{\"u}ller, Klaus-Robert and Samek, Wojciech},
  journal={IEEE transactions on neural networks and learning systems},
  volume={32},
  number={8},
  pages={3710--3722},
  year={2020},
  publisher={IEEE}
}

@article{ifca,
  title={An efficient framework for clustered federated learning},
  author={Ghosh, Avishek and Chung, Jichan and Yin, Dong and Ramchandran, Kannan},
  journal={Advances in neural information processing systems},
  volume={33},
  pages={19586--19597},
  year={2020}
}

@inproceedings{fedscope,
  title={Federatedscope-llm: A comprehensive package for fine-tuning large language models in federated learning},
  author={Kuang, Weirui and Qian, Bingchen and Li, Zitao and Chen, Daoyuan and Gao, Dawei and Pan, Xuchen and Xie, Yuexiang and Li, Yaliang and Ding, Bolin and Zhou, Jingren},
  booktitle={Proceedings of the 30th ACM SIGKDD Conference on Knowledge Discovery and Data Mining},
  pages={5260--5271},
  year={2024}
}

@article{survey_pfl,
  title={Towards personalized federated learning},
  author={Tan, Alysa Ziying and Yu, Han and Cui, Lizhen and Yang, Qiang},
  journal={IEEE transactions on neural networks and learning systems},
  volume={34},
  number={12},
  pages={9587--9603},
  year={2022},
  publisher={IEEE}
}

@INPROCEEDINGS{fedsccs,
  author={Talasso, Gabriel U. and de Souza, Allan M. and Bittencourt, Luiz F. and Cerqueira, Eduardo and Loureiro, Antonio A. F. and Villas, Leandro A.},
  booktitle={ICC 2024 - IEEE International Conference on Communications}, 
  title={FedSCCS: Hierarchical Clustering with Multiple Models for Federated Learning}, 
  year={2024},
  volume={},
  number={},
  pages={3280-3285},
  doi={10.1109/ICC51166.2024.10622346}}

@article{ffa-lora,
  title={Improving lora in privacy-preserving federated learning},
  author={Sun, Youbang and Li, Zitao and Li, Yaliang and Ding, Bolin},
  journal={arXiv preprint arXiv:2403.12313},
  year={2024}
}

@article{feddpa,
  title={Dual-personalizing adapter for federated foundation models},
  author={Long, Guodong and Shen, Tao and Jiang, Jing and Blumenstein, Michael and others},
  journal={Advances in Neural Information Processing Systems},
  volume={37},
  pages={39409--39433},
  year={2024}
}

@article{fedsa,
  title={Selective aggregation for low-rank adaptation in federated learning},
  author={Guo, Pengxin and Zeng, Shuang and Wang, Yanran and Fan, Huijie and Wang, Feifei and Qu, Liangqiong},
  journal={arXiv preprint arXiv:2410.01463},
  year={2024}
}

@article{peft,
  title={Parameter-efficient fine-tuning methods for pretrained language models: A critical review and assessment},
  author={Xu, Lingling and Xie, Haoran and Qin, Si-Zhao Joe and Tao, Xiaohui and Wang, Fu Lee},
  journal={arXiv preprint arXiv:2312.12148},
  year={2023}
}

@article{hydralora,
  title={Hydralora: An asymmetric lora architecture for efficient fine-tuning},
  author={Tian, Chunlin and Shi, Zhan and Guo, Zhijiang and Li, Li and Xu, Cheng-Zhong},
  journal={Advances in Neural Information Processing Systems},
  volume={37},
  pages={9565--9584},
  year={2024}
}

@article{FedFM,
  title={When foundation model meets federated learning: Motivations, challenges, and future directions},
  author={Zhuang, Weiming and Chen, Chen and Lyu, Lingjuan},
  journal={arXiv preprint arXiv:2306.15546},
  year={2023}
}

@article{flan,
  title={Finetuned language models are zero-shot learners},
  author={Wei, Jason and Bosma, Maarten and Zhao, Vincent Y and Guu, Kelvin and Yu, Adams Wei and Lester, Brian and Du, Nan and Dai, Andrew M and Le, Quoc V},
  journal={arXiv preprint arXiv:2109.01652},
  year={2021}
}

@article{flower,
  title={Flower: A friendly federated learning research framework},
  author={Beutel, Daniel J and Topal, Taner and Mathur, Akhil and Qiu, Xinchi and Fernandez-Marques, Javier and Gao, Yan and Sani, Lorenzo and Li, Kwing Hei and Parcollet, Titouan and De Gusm{\~a}o, Pedro Porto Buarque and others},
  journal={arXiv preprint arXiv:2007.14390},
  year={2020}
}

@InProceedings{pmlr-v54-mcmahan17a,
  title = 	 {{Communication-Efficient Learning of Deep Networks from Decentralized Data}},
  author = 	 {McMahan, Brendan and Moore, Eider and Ramage, Daniel and Hampson, Seth and Arcas, Blaise Aguera y},
  booktitle = 	 {Proceedings of the 20th International Conference on Artificial Intelligence and Statistics},
  pages = 	 {1273--1282},
  year = 	 {2017},
  editor = 	 {Singh, Aarti and Zhu, Jerry},
  volume = 	 {54},
  series = 	 {Proceedings of Machine Learning Research},
  month = 	 {20--22 Apr},
  publisher =    {PMLR},
  url = 	 {https://proceedings.mlr.press/v54/mcmahan17a.html}
}

@inproceedings{
sani2025photon,
title={Photon: Federated {LLM} Pre-Training},
author={Lorenzo Sani and Alex Iacob and Zeyu Cao and Royson Lee and Bill Marino and Yan Gao and Wanru Zhao and Dongqi Cai and Zexi Li and Xinchi Qiu and Nicholas D. Lane},
booktitle={Eighth Conference on Machine Learning and Systems},
year={2025},
url={https://openreview.net/forum?id=AQgYcfg5EI}
}

@article{crawshaw2020multi,
  title={Multi-task learning with deep neural networks: A survey},
  author={Crawshaw, Michael},
  journal={arXiv preprint arXiv:2009.09796},
  year={2020}
}

@article{smith2017federated,
  title={Federated multi-task learning},
  author={Smith, Virginia and Chiang, Chao-Kai and Sanjabi, Maziar and Talwalkar, Ameet S},
  journal={Advances in neural information processing systems},
  volume={30},
  year={2017}
}

@inproceedings{fedmllm,
  title={Leveraging Federated Learning for Multilingual and Private Language Models via Model Clustering},
  author={Talasso, Gabriel U and de Souza, Allan M and Gonzalez, Luis FG and Cerqueira, Eduardo and Loureiro, Antonio AF and Villas, Leandro A},
  booktitle={2025 3rd International Conference on Federated Learning Technologies and Applications (FLTA)},
  pages={25--32},
  year={2025},
  organization={IEEE}
}

@article{llama,
  title={The llama 3 herd of models},
  author={Grattafiori, Aaron and Dubey, Abhimanyu and Jauhri, Abhinav and Pandey, Abhinav and Kadian, Abhishek and Al-Dahle, Ahmad and Letman, Aiesha and Mathur, Akhil and Schelten, Alan and Vaughan, Alex and others},
  journal={arXiv preprint arXiv:2407.21783},
  year={2024}
}

\end{document}